\documentclass[11pt,a4paper]{article}
\usepackage[hyperref]{naaclhlt2018}
\usepackage{times}
\usepackage{latexsym}

\usepackage{url}

\usepackage{graphicx}
\usepackage{booktabs}
\usepackage{tabularx}
\usepackage{amsmath}

\usepackage{csquotes}

\aclfinalcopy 

\title{Zero-shot Sequence Labeling: \\Transferring Knowledge from Sentences to Tokens}

\author{Marek Rei \\
  The ALTA Institute \\
  Computer Laboratory \\
  University of Cambridge \\
  United Kingdom\\
  {\tt marek.rei@cl.cam.ac.uk} \\\And
  Anders S{\o}gaard \\
  CoAStaL DIKU\\
  Department of Computer Science\\
  University of Copenhagen\\
  Denmark \\
  {\tt soegaard@di.ku.dk} \\}

\date{}

\begin{document}
\maketitle
\begin{abstract}
Can attention- or gradient-based visualization techniques be used to infer token-level labels for binary sequence tagging problems, using networks trained only on sentence-level labels?
We construct a neural network architecture based on soft attention, train it as a binary sentence classifier and evaluate against token-level annotation on four different datasets.
Inferring token labels from a network provides a method for quantitatively evaluating what the model is learning, along with generating useful feedback in assistance systems.
Our results indicate that attention-based methods are able to predict token-level labels more accurately, compared to gradient-based methods, sometimes even rivaling the supervised oracle network. 
\end{abstract}

\section{Introduction}

Sequence labeling is a structured prediction task where systems need to assign the correct label to every token in the input sequence. 
Many NLP tasks, including part-of-speech tagging, named entity recognition, chunking, and error detection, are often formulated as variations of sequence labeling.
Recent state-of-the-art models make use of bidirectional LSTM architectures \cite{Irsoy2014a}, character-based representations \cite{Lample2016}, and additional external features \cite{Peters2017}.
Optimization of these models requires appropriate training data where individual tokens are manually labeled, which can be time-consuming and expensive to obtain for each different task, domain and target language.

In this paper, we investigate the task of performing sequence labeling without having access to any training data with token-level annotation.
Instead of training the model directly to predict the label for each token, the model is optimized using a sentence-level objective and a modified version of the attention mechanism is then used to infer labels for individual words.

While this approach is not expected to outperform a fully supervised sequence labeling method, it opens possibilities for making use of text classification datasets where collecting token-level annotation is not possible or cost-effective.

Inferring token-level labels from a text classification network also provides a method for analyzing and interpreting the model.
Previous work has used attention weights to visualize the focus of neural models in the input data.
However, these analyses have largely been qualitative examinations, looking at only a few examples from the datasets.
By formulating the task as a zero-shot labeling problem, we can provide quantitative evaluations of what the model is learning and where it is focusing.
This will allow us to measure whether the features that the model is learning actually match our intuition, provide informative feedback to end-users, and guide our development of future model architectures.


\section{Network Architecture}
\label{sec:arch}

The main system takes as input a sentence, separated into tokens, and outputs a binary prediction as the label of the sentence. 
We use a bidirectional LSTM \cite{Hochreiter1997} architecture for sentence classification, with dynamic attention over words for constructing the sentence representations. Related architectures have been successful for machine translation \cite{Bahdanau2015}, sentence summarization \cite{Rush2014}, entailment detection \cite{Rockt2015}, and error correction \cite{Ji2017}. In this work, we modify the attention mechanism and training objective in order to make the resulting network suitable for also inferring binary token labels, while still performing well as a sentence classifier.

Figure \ref{fig:network} contains a diagram of the network architecture. The tokens are first mapped to a sequence of word representations $[w_1, w_2, w_3, ..., w_N]$, which are constructed as a combination of regular word embeddings and character-based representations, following \newcite{Lample2016}. 
These word representations are given as input to a bidirectional LSTM which iteratively passes through the sentence in both directions. Hidden representations from each direction are concatenated at every token position, resulting in vectors $h_i$ that are focused on a specific word but take into account the context on both sides of that word. We also include a transformation with $tanh$ activation, which helps map the information from both directions into a joint feature-space:

\begin{figure*}[t]
	\centering
	\includegraphics[width=0.9\linewidth]{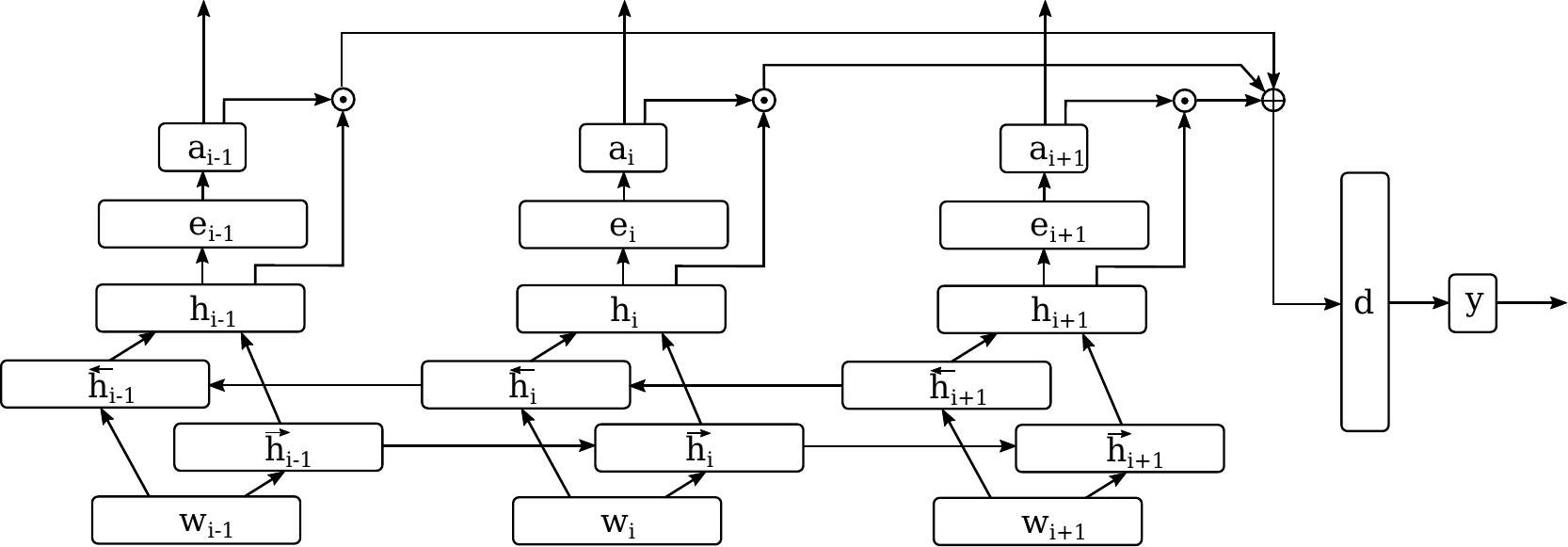}
	\caption{The neural network architecture for zero-shot sequence labeling. The soft attention values $a_i$ are used for weighting hidden representations $h_i$ as well as providing a binary label for each token. The network is only optimized through the sentence classification objective, predicting the sentence-level label $y$.}
	\label{fig:network}
\end{figure*}

\begin{equation}
\overrightarrow{h_i} = LSTM(w_i, \overrightarrow{h_{i-1}}) \\
\end{equation}
\begin{equation}
\overleftarrow{h_i} = LSTM(w_i, \overleftarrow{h_{i+1}})
\end{equation}
\vspace{-0.5cm}
\begin{equation}
\begin{aligned}[c]
\widetilde{h_i} = [\overrightarrow{h_i};\overleftarrow{h_i}] & & &  & h_i = tanh(W_h \widetilde{h_i} + b_h)\\
\end{aligned}
\label{eq:h}
\end{equation}

\noindent where $W_h$ is a parameter matrix and $b_h$ is a parameter vector, optimized during training.

Next, we include an attention mechanism that allows the network to dynamically control how much each word position contributes to the combined representation.
In most attention-based systems, the attention amount is calculated in reference to some external information. For example, in machine translation the attention values are found based on a representation of the output that has already been generated \cite{Bahdanau2015}; in question answering, the attention weights are calculated in reference to the input question \cite{Hermann2015}. In our task there is no external information to be used, therefore we predict the attention values directly based on $h_i$, by passing it through a separate feedforward layer:

\begin{equation}
e_i = tanh(W_e h_i + b_e)
\end{equation}
\begin{equation}
\widetilde{e_i} = W_{\widetilde{e}} e_i + b_{\widetilde{e}}
\end{equation}

\noindent where $W_{\widetilde{e}}$, $b_{\widetilde{e}}$, $W_e$ and $b_e$ are trainable parameters and $\widetilde{e_i}$ results in a single scalar value.
This method is equivalent to calculating the attention weights in reference to a fixed weight vector, which is optimized during training.
\newcite{Shen2016} proposed an architecture for dialogue act detection where the attention values are found based on a separate set of word embeddings. We found that the method described above was consistently equivalent or better in development experiments, while requiring a smaller number of parameters.

The values of $\widetilde{e_i}$ are unrestricted and should be normalized before using them for attention, to avoid sentences of different length having representations of different magnitude. The common approach is to use an exponential function to transform the value, and then normalize by the sum of all values in the sentence:

\begin{equation}
a_i = \frac{exp(\widetilde{e_i})}{\sum_{k=1}^N exp(\widetilde{e_k})}
\label{eq:exp}
\end{equation}

The value $a_i$ is now in a range $0 \leq a_i \leq 1$ and higher values indicate that the word at position $i$ is more important for predicting the sentence class. 
The network learns to predict informative values for $a_i$ based only on the sentence objective, without receiving token-level supervision. Therefore, we can use these attention values at each token in order to infer an unsupervised sequence labeling output.

The method in Equation \ref{eq:exp} is well-suited for applications such as machine translation -- the exponential function encourages the attention to prioritize only one word in the sentence, resulting in a word-word alignment. 
However, the same function is less suitable for our task of unsupervised sequence labeling, as there is no reason to assume that exactly one word has a positive label.
An input sentence can contain more than one tagged token, or it can contain no tokens of interest, and this should be reflected in the predictions.

Instead of the exponential function, we make use of the logistic function $\sigma$ for calculating soft attention weights:

\begin{equation}
\begin{aligned}[c]
\widetilde{a_i} = \sigma(\widetilde{e_i}) & & &  & a_i = \frac{\widetilde{a_i}}{\sum_{k=1}^N \widetilde{a_k}}\\
\end{aligned}
\end{equation}

\noindent where each $\widetilde{a_i}$ has an individual value in the range $0 \leq \widetilde{a_i} \leq 1$ and $a_i$ is normalized to sum up to $1$ over all values in the sentence. 
The normalized weights $a_i$ are used for combining the context-conditioned hidden representations from Equation \ref{eq:h} into a single sentence representation:

\begin{equation}
c = \sum_{i=1}^{N} a_i h_i
\end{equation}

\noindent In addition, we can use the pre-normalization value $\widetilde{a_i}$ as a score for sequence labeling, with a natural decision boundary of $0.5$ -- higher values indicate that the token at position $i$ is important and should be labeled positive, whereas lower values suggest the token is largely ignored for sentence classification and can receive a negative label. Attention weights with sigmoid activation have been shown to also improve performance on classification tasks \cite{Shen2016}, which indicates that this architecture has the benefit of being both accurate and interpretable on the token level.

Finally, we pass the sentence representation $c$ through a feedforward layer and predict a binary label for the overall sentence:

\begin{equation}
d = tanh(W_d c + b_d)
\end{equation}

\begin{equation}
y = \sigma(W_y d + b_y)
\end{equation}

\noindent where $d$ is a sentence vector and $y$ is a single value between $0 \leq y \leq 1$, with values higher than $0.5$ indicating a positive class and lower values indicating a negative prediction.

In order to optimize the model, we use several different loss functions. The first is the squared loss which optimizes the sentence-level score prediction to match the gold label in the annotation:

\begin{equation}
L_1 = \sum_j (y^{(j)} - \widetilde{y}^{(j)})^2
\label{eq:l1}
\end{equation}

\noindent where $y^{(j)}$ is the predicted score for the $j$-th sentence, and $\widetilde{y}^{(j)}$ is the true binary label $(0, 1)$ for the $j$-th sentence.

In addition, we want to encourage the model to learn high-quality token-level labels as part of the attention weights. While the model does not have access to token-level annotation during training, there are two constraints that we can take advantage of:

\begin{enumerate}
\item Only some, but not all, tokens in the sentence can have a positive label.
\item There are positive tokens in a sentence only if the overall sentence is positive.
\end{enumerate}

We can then construct loss functions that encourage the model to optimize for these constraints:

\begin{equation}
L_2 = \sum_j (min_i(\widetilde{a_i}) - 0)^2
\label{eq:min}
\end{equation}

\begin{equation}
L_3 = \sum_j (max_i(\widetilde{a_i}) - \widetilde{y}^{(j)})^2
\label{eq:max}
\end{equation}

\noindent where $min_i(\widetilde{a_i})$ is the minimum value of all the attention weights in the sentence and $max_i(\widetilde{a_i})$ is the corresponding maximum value.
Equation \ref{eq:min} optimizes the minimum unnormalized attention weight in a sentence to be 0, satisfying the constraint that all tokens in a sentence should not have a positive token-level label.
Equation \ref{eq:max} then optimizes for the maximum unnormalized attention weight in a sentence to be equal to the gold label for that sentence, which is either $0$ or $1$, incentivizing the network to only assign large attention weights to tokens in positive sentences.
These objectives do not provide the model with additional information, but serve to push the attention scores to a range that is suitable for binary classification.

We combine all of these loss objectives together for the main optimization function:

\begin{equation}
L = L_1 + \gamma (L_2 + L_3)
\end{equation}

\noindent where $\gamma$ is used to control the importance of the auxiliary objectives.

\section{Alternative Methods}

We compare the attention-based system for inferring sequence labeling with 3 alternative methods.

\subsection{Labeling Through Backpropagation}
\label{sec:backprop}

We experiment with an alternative method for inducing token-level labels, based on visualization methods using gradient analysis.
Research in computer vision has shown that interpretable visualizations of convolutional networks can be obtained by analyzing the gradient after a single backpropagation pass through the network \cite{Zeiler2014}. 
\newcite{Denil2014} extended this approach to natural language processing, in order to find and visualize the most important sentences in a text.
Recent work has also used the gradient-based approach for visualizing the decisions of text classification models on the token level \cite{Li2016,Alikaniotis2016}.
In this section we propose an adaptation that can be used for sequence labeling tasks.

We first perform a forward pass through the network and calculate the predicted sentence-level score $y$. Next, we define a pseudo-label $y^* = 0$, regardless of the true label of the sentence. We then calculate the gradient of the word representation $w_i$ with respect to the loss function using this pseudo-label:

\begin{equation}
g_i = \frac{\partial L_1}{\partial w_i} \Bigr|_{\substack{(y^*,y)}}
\end{equation}

\noindent where $L_1$ is the squared loss function from Equation \ref{eq:l1}. The magnitude of $g_i$, $|g_i|$ can now be used as an indicator of how important that word is for the positive class. The intuition behind this approach is that the magnitude of the gradient indicates which individual words need to be changed the most in order to make the overall label of the sentence negative. These are the words that are contributing most towards the positive class and should be labeled as such individually.

An obstacle in using this score for sequence labeling comes from the fact that there is no natural decision boundary between the two classes. The magnitude of the gradient is not constrained to a specific range and can vary quite a bit depending on the sentence length and the predicted sentence-level score.
In order to map this magnitude to a decision, we analyze the distribution of magnitudes in a sentence. Intuitively, we want to detect outliers -- scores that are larger than expected. Therefore, we map all the magnitudes in a sentence to a Gaussian distribution and set the decision boundary at $1.5$ standard deviations. Any word that has a gradient magnitude higher than that will be tagged with a positive class for sequence labeling. If all the magnitudes in a sentence are very similar, none of them will cross this threshold and therefore all words will be labeled as negative.

We calculate the gradient magnitude using the same network architecture as described in Section \ref{sec:arch}, at word representation $w_i$ after the character-based features have been included. The attention-based architecture is not necessary for this method, therefore we also report results using a more traditional bidirectional LSTM, concatenating the last hidden states from both directions and using the result as a sentence representation for the main objective.

\subsection{Relative Frequency Baseline}
\label{sec:relfreq}

The system for producing token-level predictions based on sentence-level training data does not necessarily need to be a neural network. As the initial experiment, we trained a Naive Bayes classifier with n-gram features on the annotated sentences and then used it to predict a label only based on a window around the target word. However, this did not produce reliable results -- since the classifier is trained on full sentences, the distribution of features is very different and does not apply to a window of only a few words.

Instead, we calculate the relative frequency of a feature occurring in a positive sentence, normalized by the overall frequency of the feature, and calculate the geometric average over all features that contain a specific word:

\begin{equation}
r_k = \frac{c(X_k=1, Y=1)}{\sum_{z \in (0,1)} c(X_k=1, Y=z)}  
\end{equation}
\begin{equation}
score_i = \sqrt[\leftroot{-2}\uproot{4}|F_i|]{ \prod_{k \in F_i} r_k }
\end{equation}

\noindent where $c(X_k=1, Y=1)$ is the number of times feature $k$ is present in a sentence with a positive label, $F_i$ is the set of n-gram features present in the sentence that involve the $i$-th word in the sentence, and $score_i$ is the token-level score for the $i$-th token in the sentence.
We used unigram, bigram and trigram features, with extra special tokens to mark the beginning and end of a sentence.

This method will assign a high score to tokens or token sequences that appear more often in sentences which receive a positive label. While it is not able to capture long-distance context, it can memorize important keywords from the training data, such as modal verbs for uncertainty detection or common spelling errors for grammatical error detection.

\subsection{Supervised Sequence Labeling}
\label{sec:supervised}

Finally, we also report the performance of a supervised sequence labeling model on the same tasks. This serves as an indicator of an upper bound for a given dataset -- how well the system is able to detect relevant tokens when directly optimized for sequence labeling and provided with token-level annotation.

We construct a bidirectional LSTM tagger, following the architectures from \newcite{Irsoy2014a},  \newcite{Lample2016} and \newcite{Rei2017}. Character-based representations are concatenated with word embeddings, passed through a bidirectional LSTM, and the hidden states from both direction are concatenated. Based on this, a probability distribution over the possible labels is predicted and the most probable label is chosen for each word. While \newcite{Lample2016} used a CRF on top of the network, we exclude it here as the token-level scores coming from that network do not necessarily reflect the individual labels, since the best label sequence is chosen globally based on the combined sentence-level score.
The supervised model is optimized by minimizing cross-entropy, training directly on the token-level annotation.

\begin{table*}[t]
\setlength\tabcolsep{5.2pt}
\begin{tabular}{r|ccccc|ccccc} \toprule
 & \multicolumn{5}{c|}{CoNLL 2010} & \multicolumn{5}{c}{FCE} \\
 & Sent $F_1$ & MAP & P & R & $F_1$ & Sent $F_1$ & MAP & P & R & $F_1$ \\ \midrule
Supervised & - & 96.54 & 78.92 & 79.41 & 79.08 & - & 59.13 & 49.15 & 26.96 & 34.76 \\ \midrule
Relative freq & - & 81.78 & 15.94 & \textbf{79.98} & 26.59 & - & 37.75 & 14.37 & \textbf{86.36} & 24.63 \\
\small{LSTM-LAST-BP} & 84.42 & 77.90 & 7.16 & 66.64 & 12.92 & 85.10 & 46.12 & \textbf{29.49} & 16.07 & 20.80 \\
\small{LSTM-ATTN-BP} & \textbf{84.94} & 80.38 & 9.13 & 71.42 & 16.18 & \textbf{85.14} & 44.52 & 27.62 & 17.81 & 21.65 \\
\small{LSTM-ATTN-SW} & \textbf{84.94} & \textbf{87.86} & \textbf{77.48} & 69.54 & \textbf{73.26} & \textbf{85.14} & \textbf{47.79} & 28.04 & 29.91 & \textbf{28.27} \\ \bottomrule
\end{tabular}
\caption{Results for different system configurations on the CoNLL 2010 and FCE datasets. Reporting sentence-level $F_1$, token-level Mean Average Precision (MAP), and token-level precision/recall/$F_1$.}
\label{tab:results1}
\end{table*}

\section{Datasets}

We evaluate the performance of zero-shot sequence labeling on 3 different datasets. In each experiment, the models are trained using only sentence-level annotation and then evaluated based on token-level annotation.

\subsection{CoNLL 2010 Uncertainty Detection}

The CoNLL 2010 shared task \cite{Farkas2010} investigated the detection of uncertainty in natural language texts.
The use of uncertain language (also known as hedging) is a common tool in scientific writing, allowing scientists to guide research beyond the evidence without overstating what follows from their work. \newcite{Vincze2008} showed that 19.44\% of  sentences in the biomedical papers of the BioScope corpus contain hedge cues.
Automatic detection of these cues is important for downstream tasks such as information extraction and literature curation, as typically only definite information should be extracted and curated.

The dataset is annotated for both hedge cues (keywords indicating uncertainty) and scopes (the area of the sentence where the uncertainty applies). The cues are not limited to single tokens, and can also consist of several disjoint tokens (for example, \textit{"either ... or ..."}).
An example sentence from the dataset, with bold font indicating the hedge cue and curly brackets marking the scope of uncertainty:

\begin{displayquote}
Although IL-1 has been reported to contribute to Th17 differentiation in mouse and man, it remains to be determined \{\textbf{whether} therapeutic targeting of IL-1 will substantially affect IL-17 in RA\}.
\end{displayquote}

The first subtask in CoNLL 2010 was to detect any uncertainty in a sentence by predicting a binary label.
The second subtask required the detection of all the individual cue tokens and the resolution of their scope.
In our experiments, we train the system to detect sentence-level uncertainty, use the architecture to infer the token-level labeling and evaluate the latter on the task of detecting uncertainty cues. 
Since the cues are defined as keywords that indicate uncertainty, we would expect the network to detect and prioritize attention on these tokens.
We use the train/test data from the second task, which contains the token-level annotation needed for evaluation, and randomly separate 10\% of the training data for development.

\subsection{FCE Error Detection}

Error detection is the task of identifying tokens which need to be edited in order to produce a grammatically correct sentence. The task has numerous applications for writing improvement and assessment, and recent work has focused on error detection as a supervised sequence labeling task \cite{Rei2016,Kaneko2017,Rei2017}. 

Error detection can also be performed on the sentence level -- detecting whether the sentence needs to be edited or not. \newcite{Andersen2013} described a practical tutoring system that provides sentence-level feedback to language learners. 
The 2016 shared task on Automated Evaluation of Scientific Writing \cite{Daudaravicius2016} also required participants to return binary predictions on whether the input sentence needs to be corrected. 

We evaluate our system on the First Certificate in English (FCE, \newcite{Yannakoudakis2011}) dataset, containing error-annotated short essays written by language learners. 
While the original corpus is focused on aligned corrections, \newcite{Rei2016} converted the dataset to a sequence labeling format, which we make use of here.
An example from the dataset, with bold font indicating tokens that have been annotated as incorrect given the context:

\begin{displayquote}
When the show started the person who was acting \textbf{it} was not Danny Brook and \textbf{he seemed not} to be an actor.
\end{displayquote}

We train the network as a sentence-level error detection system, returning a binary label and a confidence score, and also evaluate how accurately it is able to recover the locations of individual errors on the token level.

\subsection{SemEval Sentiment Detection in Twitter}

SemEval has been running a series of popular shared tasks on sentiment analysis in text from social media \cite{Nakov2013,Rosenthal2014,Rosenthal2015}.
The competitions have included various subtasks, of which we are interested in two: Task A required the polarity detection of individual phrases in a tweet, and Task B required sentiment detection of the tweet as a whole.
A single tweet could contain both positive and negative phrases, regardless of its overall polarity, and was therefore separately annotated on the tweet level.

In the following example from the dataset, negative phrases are indicated with a bold font and positive phrases are marked with italics, whereas the overall sentiment of the tweet is annotated as negative:

\begin{displayquote}
They may \textit{have} a \textit{SuperBowl} in Dallas, but Dallas \textbf{ain't winning} a SuperBowl. \textbf{Not with that} quarterback and owner. @S4NYC @RasmussenPoll
\end{displayquote}

Sentiment analysis is a three-way task, as the system needs to differentiate between positive, negative and neutral sentences. Our system relies on a binary signal, therefore we convert this dataset into two binary tasks -- one aims to detect positive sentiment, the other focuses on negative sentiment.
We train the system as a sentiment classifier, using the tweet-level annotation, and then evaluate the system on recovering the individual positive or negative tokens.
We use the train/dev/test splits of the original SemEval 2013 Twitter dataset, which contains phrase-level sentiment annotation.

\begin{table*}[t]
\setlength\tabcolsep{5.2pt}
\begin{tabular}{r|ccccc|ccccc} \toprule
 & \multicolumn{5}{c|}{SemEval Negative} & \multicolumn{5}{c}{SemEval Positive} \\
 & Sent $F_1$ & MAP & P & R & $F_1$ & Sent $F_1$ & MAP & P & R & $F_1$ \\ \midrule
Supervised & - & 67.70 & 31.79 & 44.66 & 37.02 & - & 67.41 & 36.27 & 50.71 & 42.24 \\ \midrule
Relative freq & - & 44.15 & 17.39 & 15.67 & 16.48 & - & 47.64 & 13.39 & \textbf{54.69} & 21.51 \\
\small{LSTM-LAST-BP} & 53.65 & 43.02 & 8.33 & 28.41 & 12.88 & 70.83 & 49.06 & 17.66 & 35.06 & 23.48 \\
\small{LSTM-ATTN-BP} & \textbf{55.83} & 50.96 & 11.55 & \textbf{31.54} & 16.90 & \textbf{71.26} & 53.89 & 23.45 & 34.53 & 27.92 \\
\small{LSTM-ATTN-SW} & \textbf{55.83} & \textbf{54.37} & \textbf{29.41} & 14.40 & \textbf{19.23} & \textbf{71.26} & \textbf{56.45} & \textbf{37.19} & 25.96 & \textbf{30.45} \\ \bottomrule
\end{tabular}
\caption{Results for different system configurations on the SemEval Twitter sentiment dataset, separated into positive and negative sentiment detection. Reporting sentence-level $F_1$, token-level Mean Average Precision (MAP), and token-level precision/recall/$F_1$.}
\label{tab:results2}
\end{table*}

\section{Implementation Details}

During pre-processing, tokens are lowercased while the character-level component still retains access to the capitalization information.
Word embeddings were set to size 300, pre-loaded from publicly available Glove \cite{Pennington} embeddings and fine-tuned during training.
Character embeddings were set to size 100. 
The recurrent layers in the character-level component have hidden layers of size $100$; the hidden layers $\overrightarrow{h_i}$ and $\overleftarrow{h_i}$ are size 300. 
The hidden combined representation $h_i$ was set to size 200, and the attention weight layer $e_i$ was set to size 100.
Parameter $\gamma$ was set to $0.01$ based on development experiments.

The model was implemented using Tensorflow \cite{Abadi2016}.
The network weights were randomly initialized using the uniform Glorot initialization method \cite{Glorot2010} and optimization was performed using AdaDelta \cite{Zeiler2012} with learning rate $1.0$.
Dropout \cite{Srivastava2014a} with probability $0.5$ was applied to word representations $w_i$ and the composed representations $h_i$ after the LSTMs.
The training was performed in batches of 32 sentences. 
Sentence-level performance was observed on the development data and the training was stopped if performance did not improve for 7 epochs. 
The best overall model on the development set was then used to report performance on the test data, both for sentence classification and sequence labeling.
In order to avoid random outliers, we performed each experiment with 5 random seeds and report here the averaged results.

The code used for performing these experiments is made available online.\footnote{http://www.marekrei.com/projects/mltagger}

\begin{figure*}[t]
	\centering
\includegraphics[width=1.0\linewidth]{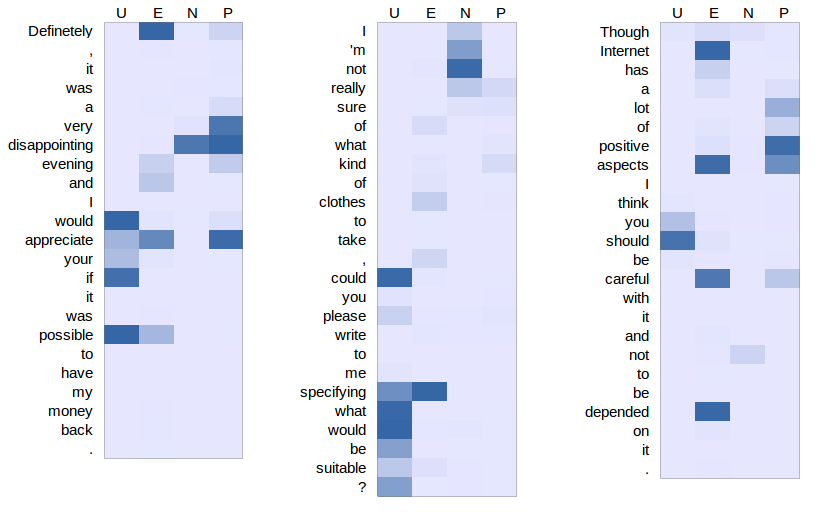}
	\caption{Example output from each of the zero-shot sequence labeling models, trained on 4 different tasks. \textbf{U}: uncertainty detection, \textbf{E}: error detection, \textbf{N}: negative sentiment detection, \textbf{P}: positive sentiment detection. Darker blue indicates higher predicted values.}
    \label{fig:examples}
\end{figure*}

\section{Evaluation}

Results for the experiments are presented in Tables \ref{tab:results1} and \ref{tab:results2}. 
We first report the sentence-level F-measure in order to evaluate the performance on the general text classification objective.
Next, we report the Mean Average Precision (MAP) at returning the active/positive tokens. This measure rewards systems that assign higher scores to positive tokens as opposed to negative ones, evaluating this as a ranking problem. It disregards a specific classification threshold and therefore provides a more fair evaluation towards systems that could be improved simply by choosing a different decision boundary. 
Finally, we also report token-level precision, recall and F-measure for evaluating the accuracy of this model as a sequence labeler.\footnote{The CoNLL 2010 shared task on uncertainty detection comes with an official scorer which requires additional steps and the detection of both cues and scopes, whereas the binary labels from the zero-shot systems are not directly applicable to this format. Similarly, error detection is commonly evaluated using $F_{0.5}$, which is motivated by end-user experience, but in this case we wish to specifically measure the tagging accuracy. Therefore we use the regular $F_1$ score as the main evaluation metric for both of these tasks.}

We report five different system configurations: 
\textbf{Relative freq} is the n-gram based approach described in Section \ref{sec:relfreq}.
\textbf{Supervised} is the fully supervised sequence labeling system described in Section \ref{sec:supervised}.
\textbf{\small{LSTM-LAST-BP}} is using the last hidden states from the word-level LSTMs for constructing a sentence representation, and the backpropagation-based method from Section \ref{sec:backprop} for inducing token labels. \textbf{\small{LSTM-ATTN-BP}} is using the attention-based network architecture together with the backpropagation-based labeling method. \textbf{\small{LSTM-ATTN-SW}} is the method described in Section \ref{sec:arch}, using soft attention weights for sequence labeling and additional objectives for optimizing the network.

The method using attention weights achieves the best performance on all datasets, compared to other methods not using token-level supervision.
On the CoNLL 2010 uncertainty detection dataset the system reaches 73.26\% F-score, which is 93\% of the supervised upper bound. 
The alternative methods using backpropagation and relative frequency achieve high recall values, but comparatively lower precision. 
On the FCE dataset, the F-score is considerably lower at 28.27\% -- this is due to the difficulty of the task and the supervised system also achieves only 34.76\%.
The attention-based system outperforms the alternatives on both of the SemEval evaluations. The task of detecting sentiment on the token level is quite difficult overall as many annotations are context-specific and require prior knowledge. For example, in order to correctly label the phrase \textit{"have Superbowl"} as positive, the system will need to understand that organizing the Superbowl is a positive event for the city.

Performance on the sentence-level classification task is similar for the different architectures on the CoNLL 2010 and FCE datasets, whereas the composition method based on attention obtains an advantage on the SemEval datasets.
Since the latter architecture achieves competitive performance and also allows for attention-based token labeling, it appears to be the better choice.
Analysis of the token-level MAP scores shows that the attention-based sequence labeling model achieves the best performance even when ignoring classification thresholds and evaluating the task through ranking.

Figure \ref{fig:examples} contains example outputs from the attention-based models, trained on each of the four datasets.
In the first example, the uncertainty detector correctly picks up \textit{"would appreciate if"} and \textit{"possible"}, and the error detection model focuses most on the misspelling \textit{"Definetely"}.
Both the positive and negative sentiment models have assigned a high weight to the word \textit{"disappointing"}, which is something we observed in other examples as well. 
The system will learn to focus on phrases that help it detect positive sentiment, but the presence of negative sentiment provides implicit evidence that the overall label is likely not positive. This is a by-product of the 3-way classification task and future work could investigate methods for extending zero-shot classification to better match this requirement.

In the second example, the system correctly labels the phrase \textit{"what would be suitable?"} as uncertain, and part of the phrase \textit{"I'm not really sure"} as negative. It also labels \textit{"specifying"} as an error, possibly expecting a comma before it.
In the third example, the error detection model labels \textit{"Internet"} for the missing determiner, but also captures a more difficult error in \textit{"depended"}, which is an incorrect form of the word given the context.

\section{Conclusion}

We investigated the task of performing sequence labeling without having access to any training data with token-level annotation.
The proposed model is optimized as a sentence classifier and an attention mechanism is used for both composing the sentence representations and inferring individual token labels.
Several alternative models were compared on three tasks -- uncertainty detection, error detection and sentiment detection.

Experiments showed that the zero-shot labeling system based on attention weights achieved the best performance on all tasks. The model is able to automatically focus on the most salient areas of the sentence, and additional objective functions along with the soft attention mechanism encourage it to also perform well as a sequence labeler.
The zero-shot labeling task can provide a quantitative evaluation of what the model is learning, along with offering a low-cost method for creating sequence labelers for new tasks, domains and languages.

\section*{Acknowledgments}

We would like to thank the NVIDIA Corporation for the donation of the Titan GPU that was used for this research. Anders S{\o}gaard was partially funded by the ERC Starting Grant LOWLANDS No. 313695.


\bibliography{interpsent}
\bibliographystyle{acl_natbib}

\end{document}